\pdfoutput=1
\documentclass[11pt]{article}
\usepackage{hyperref}
\hypersetup{
    colorlinks,
    linkcolor={black!50!black},
    citecolor={black!50!black},
    urlcolor={black!80!black},
    pagebackref=false}
\usepackage{eamt23}
\usepackage{times}
\usepackage{xurl}
\usepackage{latexsym}
\usepackage[small,bf]{caption} 
\usepackage{float}
\usepackage{tipa}
\usepackage{xcolor}
\usepackage{tabularx}
\usepackage{tabto}
\usepackage{soul}
\setuldepth{test}

\title{Gender Neutralization for an Inclusive Machine Translation: from Theoretical Foundations to Open Challenges}

\author{
  Andrea Piergentili\textsuperscript{1,2 $\ast$},
  Dennis Fucci\textsuperscript{1,2 $\ast$},\\ 
  \textbf{  Beatrice Savoldi\textsuperscript{1}, Matteo Negri\textsuperscript{1}, Luisa Bentivogli\textsuperscript{1}}\\
  \textsuperscript{1}Fondazione Bruno Kessler \quad
  \textsuperscript{2}University of Trento\\
  {\tt \{apiergentili,dfucci,bsavoldi,negri,bentivo\}@fbk.eu}}

\date{}

\begin{document}
\maketitle
\begin{abstract}
\newcommand\blfootnote[1]{%
  \begingroup
  \renewcommand\thefootnote{}\footnote{#1}%
  \addtocounter{footnote}{-1}%
  \endgroup
}
\blfootnote{\textsuperscript{$\ast$}The authors contributed equally.}
Gender inclusivity in language technologies has become a prominent research topic.
In this study, we explore gender-neutral translation (GNT) as a form of gender inclusivity and a goal to be achieved by machine translation (MT) models, which have been found to perpetuate gender bias and discrimination.
Specifically, we focus on translation from English into Italian, 
a language pair representative of salient gender-related linguistic transfer problems. 
To define GNT, we review a selection of relevant institutional guidelines for gender-inclusive language, discuss its scenarios of use,
and examine the technical challenges of performing GNT in MT, concluding with a discussion of potential solutions to encourage advancements toward greater inclusivity in MT.
\end{abstract}

\section{Introduction}

Language technologies have become ubiquitous and play a significant role in our lives. In addition to their benefits, however, these technologies come with 
potential ethical shortcomings and risks \cite{blodgett-etal-2020-language}.
Among them is gender bias, whose presence in  machine translation (MT) is well-documented \cite{savoldi2021gender}.
Indeed, MT systems were found to systematically prefer masculine forms 
(e.g., EN \textit{The student} → IT \textit{\ul{Lo}} (M) \textit{studente})
and stereotypical gender associations in their outputs (e.g., EN \textit{The doctors and the nurses} → IT \textit{\ul{I} dottori} (M) \textit{e \ul{le} infermiere} (F)), 
thus reinforcing bias and reiterating the under-representation of specific groups \cite{savoldi2021gender}. 
As the role of gender is relevant on the social level \cite{kiesling2019language} and for each individual's construction of identity \cite{Crenshaw1991}, the biased behaviors of MT systems give rise to concerns about the consequent risks. These risks rest on the power of language to reproduce and reinforce societal asymmetries \cite{Lazar2005}, and its impact on our perception \cite{Boroditsky2003,Gygaxetal2008}. 

%
Spurred by the ever-growing demand for a gender-inclusive language, in this work we explore gender-\textit{neutral} language as a form of gender inclusivity. 
It conforms to standard and established linguistic resources that allow to avoid gendered forms (e.g., \textit{chairperson} instead of \textit{chairman}) – unlike innovative elements like neopronouns and neomorphemes, which are not considered acceptable in many contexts (see Section \ref{sec:genderlanguage}). 
%
Comprehensive inquiries on gender-neutral MT are largely absent and its implementation is a substantially uncharted territory. Such gap calls for dedicated work on methodological underpinnings, such as the definition of the objectives and an outline of the main challenges to be faced when developing gender-neutral MT systems.

In light of the foregoing, in the present work we discuss the implementation of inclusive language in MT, through the definition of a novel task for MT: \textit{gender-neutral translation} (GNT). For this purpose, we first provide a brief account of the relation between gender and language, and gender bias in MT (Section \ref{sec:background}). 
Then, we focus on English-Italian translation and start by analysing relevant guidelines 
for gender inclusivity in both languages to understand the current theoretical frameworks 
(Section \ref{sec:guidelines}). 
We chose this language pair because 
it is representative of the challenges faced by MT systems when translating across languages that express gender differently. 
This mismatch 
can result in
undesired and discriminatory phenomena, such as the misgendering of individuals or stereotyped translations. 
Thereafter, we integrate the main principles of the guidelines into the context of MT, thus outlining a set of desiderata which define the task of GNT in higher detail (Section \ref{sec:desiderata}). Finally, we discuss the open technical challenges that performing and evaluating GNT in MT entails, and examine the potential approaches to address them, thus sketching a road-map towards the implementation of GNT in MT (Section \ref{sec:challenges}).

\section{Background}
\label{sec:background}

Gender is a complex notion, which has been extensively debated across several disciplines.
In the case of language, the relationship with gender is socially relevant (%
Section \ref{sec:genderlanguage}), with an impact on both the visibility of gender groups \cite{wasserman-2009
} and the construction of personal identities \cite{gygax-2019}. 
Therefore, the appropriate use of gender expressions is critical in communicative practices, including those allowed 
by language technologies (%
Section \ref{sec:genderbias}).

\subsection{Gender in Language}
\label{sec:genderlanguage}

The concept of gender is so relevant to human experience that no language lacks expressions of femaleness or maleness altogether \cite{stahlberg2007representation}. However, languages differ in how they encode gender. English, for example, is a notional gender language: it expresses the gender of human referents mostly through personal pronouns and possessive adjectives (e.g., \textit{he/him/hers}; \textit{she/her/hers}), and lexically gendered forms (e.g., \textit{man}; \textit{woman})
. Grammatical gender languages like Italian, instead, are characterized by a system of morphosyntactic agreement, where several parts of speech beside the noun (e.g., verbs, determiners, adjectives) carry gender inflections, as in \textit{\ul{I/Le} bambin\ul{i}/bambin\ul{e} sono content\ul{i}/content\ul{e}} (EN \textit{The children are happy}). Such differences are particularly relevant in translation, especially when the
source language does not provide gender information about a referent and the target is a grammatical gender language, as in the previous example.

Regardless of cross-lingual differences, however, linguistic practices can be discriminatory when they generate a disparity in the representation of the genders based on normative and stereotypical principles.
Androcentric normativity promotes the masculine gender as the human prototype, 
encompassing the whole human experience \cite{hellinger-pauwels2007}, thus treating women as a gendered deviation from the norm. A typical manifestation of normativity in language is the masculine generic, i.e., the use of masculine forms as conceptually generic, neutral (e.g., \textit{one must watch \ul{his} language}), when referring to mixed-gender groups or when gender is unknown or unspecified. 
Stereotypes, instead, are reiterated 
in the assumption of someone’s gender through associations of professional nouns and gender (e.g., \textit{nurse} = feminine, \textit{doctor} = masculine) \cite{he2010}, fostering unfair gender paradigms
. Moreover, within binary gender linguistic frameworks, non-binary experiences are 
completely omitted from representation.


In light of this, we look at gender-inclusive language\footnote{The label “inclusive language” covers a wide range of linguistic practices aimed at avoiding discrimination and denigration on any basis (see 
\url{https://www.apa.org/about/apa/equity-diversity-inclusion/language-guidelines}). 
Such practices have also been given different labels, such as ‘neutral’ and ‘fair’. To set the object of our analysis within a larger scope of inclusivity, we hereby rely on the label \textit{gender-inclusive language}.} 
for the avoidance of 
discriminatory language. This is a form of \textit{verbal hygiene} \cite{cameron1995verbal} by which people attempt to regulate language in conformity to certain ideals, and promote linguistic policies that reflect them.
The efforts to make language fair and 
inclusive of all gender identities can be distinguished in two main approaches: \textit{i)} the introduction of innovative linguistic resources 
, and \textit{ii)} the use of gender-neutral formulations. The first approach is the result of ongoing grassroots efforts, and includes neopronouns (EN \textit{ze/zir} instead of \textit{he/she/him/his/her}), neomorphemes (ES \textit{-e/-es} instead of \textit{-o/-os} and \textit{-a/-as}), and other solutions (e.g., graphemic devices such as IT \textit{-@} instead of\textit{ -a/-o/-e/-i}
) that allow to mention referents without resorting to generic terms. 
The 
acceptance of such resources, however, is still highly debated and mostly restricted to informal communication channels like social media \cite{Comandini_2021}. Thus, speakers who wish to use a gender-inclusive language in more formal contexts can turn to the second approach, which solely relies on established gender-neutral devices of the standard language. While some languages already feature convenient gender-neutral resources, such as the well established singular \textit{they} in English,\footnote{
See the American Psychological Association's style guidelines: 
\url{https://apastyle.apa.org/style-grammar-guidelines/grammar/singular-they}
} 
speakers of other languages, such as Italian, cannot rely on similar devices. 
Then, they can resort to gender-neutralization strategies, such as the preference for epicene words, i.e. words that are not gender-marked and can be used regardless of the referent's gender (e.g., \textit{spokesperson}, as opposed to \textit{spokesman} and \textit{spokeswoman}).
Neutralization strategies range from simple word choices to complex sentence formulations without introducing innovative elements, thus being aligned with standardized forms and grammar.
Consequently, we look at gender neutralization as a viable and grammatically acceptable form of gender-inclusive language, and a more solid ground for the exploration of gender-inclusive MT.

\subsection{Gender (Bias) in Machine Translation}
\label{sec:genderbias}

Although affecting many monolingual tasks in natural language processing (NLP), gender bias comes across 
more evident in cross-lingual scenarios, such as the case of MT, where different languages can encode very different gender marking mechanisms
\cite[\textit{inter alia}]{Prates2020,savoldi2021gender}.
Most efforts to address gender bias in MT still operate in the binary perspective 
\cite[\textit{inter alia}]{vanmassenhove-etal-2018-getting,stafanovics-etal-2020-mitigating}, thus ignoring the neutral solutions.
By using gender-neutral forms, it is possible to avoid undue gendering when no information about the referents' gender is available, while also 
including all gender identities.

Recently, some works have started working on the processing of non-binary gender forms in NLP and highlighted the main challenges involved \cite{dev-etal-2021-harms,lauscher-etal-2022-welcome}. They mainly focused on standard neutral solutions for text classification \cite{attanasio-2021}, coreference resolution \cite{cao-daume-iii-2020-toward}, and natural language generation tasks, such as gender-neutral rewriting \cite{sun2021they,vanmassenhove-etal-2021-neutral,attanasio-2021}.
As regards MT, Cho et al. \shortcite{cho-etal-2019-measuring} built a benchmark with template sentences to evaluate whether gender neutrality is preserved in Korean~→~English automatic translations. 
Working on English~→~German/Spanish, Saunders \& Byrne \shortcite{saunders-byrne-2020-reducing} also created a benchmark to assess the ability of MT systems to generate neutral target sentences. 
As the considered target languages 
do not have a neutral gender, they used gender-neutral placeholders for articles and inflectional morphemes. 
Finally, specific projects dedicated to gender-inclusive translation are also arising, like GenderFairMT,\footnote{See \url{https://genderfair.univie.ac.at/index_en.html}
} 
with a focus on inclusive solutions for English → German 
MT \cite{Burtscher}.

Overall, adopting a neutral translation as a path towards gender inclusivity poses non-negligible challenges to MT. On the one hand, the complexity of implementing neutral forms comes from the inherent difficulties posed by grammatical gender languages, as also exemplified by the case study in \cite{saunders-byrne-2020-reducing}. On the other hand, the application of an inclusive language must be carefully designed not to be perceived as intrusive nor as language policing. 

In light of the foregoing, before we confront the technical challenges that arise from gender-neutralization in MT, we need to lay the groundwork for this endeavor. That is, framing the linguistic possibilities that could be adopted towards an automatic neutral translation, and identifying their suitable deployment.

\section{Framing Gender-Inclusive Language}
\label{sec:guidelines}

Looking for guidance to determine how MT systems should adopt gender-inclusive language, the MT scenario remains largely unexplored. Nonetheless, several resources intended for (human) communication are available 
and offer valuable linguistic knowledge for the understanding of gender-inclusive language and towards its adoption in MT.
Among the most influential and accessible resources, there are the guidelines produced 
by renowned institutions to address gender discrimination in language. 
We consider them ‘top-down’ approaches in language, as opposed to the ‘bottom-up’ efforts of grassroots movements. 
Institutional guidelines currently only address monolingual communication while our domain of interest is translation. However, we analyze them to collect useful inclusive linguistic strategies, which let us investigate GNT and discuss its practical implications.
More precisely, we intend to \textit{i)} explore how gender inclusivity is conceptualized within such guidelines (Section \ref{sec:conceptualization}), and \textit{ii)} gain insights concerning \textit{what} should be neutralized and \textit{how} it should be neutralized (Section \ref{sec:what-how}). 

To this aim, we selected 30 guidelines published online\footnote{Retrieved through Google queries on October 28, 2022.} by relevant institutions, equally divided between guidelines for English and Italian (see the full list of guidelines in Appendix \ref{guidelines}).
Besides prestige, we prioritized comparability: we selected guidelines by international institutions (e.g., the European Union) that published the same document in both languages, or by national institutions (e.g., universities and governmental bodies) that share a similar status across countries, thus also ensuring that the selected guidelines belong to the same textual genre.

\subsection{Conceptualization of Gender}
\label{sec:conceptualization}
Starting from \textit{\textbf{how these inclusive guidelines interpret gender}}, and hence gender-based discrimination, we find clear differences between the English and the Italian documents. While the former mostly go beyond the binary gender framework, the Italian guidelines tend to address women and men only. 
Such a difference emerges clearly in the two versions of the European Parliament’s guidelines (see documents E3, I5 in the reference list). This fundamental difference reflects different ideas of discrimination (e.g., E3: “achieving equality”, I5: “achieving equality between men and women”). This conceptual discrepancy is reflected in the suggested strategies to address discrimination at the linguistic level. For instance, the Italian guidelines provide extensive lists of feminine counterparts for traditionally masculine professional nouns (e.g., EN \textit{coordinator} as IT \textit{coordina\ul{tore}} [M] / \textit{coordina\ul{trice}} [F]). Also, they often endorse gender specification to avoid masculine generics (e.g., EN \textit{The professors}~
→~\textit{\ul{I} profess\ul{ori}} [M] \textit{e \ul{le} profess\ul{oresse}} [F]). Since such suggestions remain within a binary framework, they 
do not conform to
our gender-neutral goal, and are hence discarded in 
the following discussion.

\begin{table*}[!ht]
\small
\renewcommand{\arraystretch}{1.2}
\centering
\begin{tabularx}{450pt}{c|c|l}
\hline
\multicolumn{3}{l}{\textbf{A. Epicene synonyms}}\\
\hline
EN & E5 & \textbf{\color{red}\textit{Chairman}} → \textbf{\color{teal}\ul{Chair(person)}} \\
IT & I3 &\textbf{\color{red}\textit{Professore}} [Professor] → \textbf{\color{teal}\ul{Docente}} [Teacher]  \\
\hline
\multicolumn{3}{l}{\textbf{B. Pluralization (towards generic or epicene forms)}}\\
\hline
EN & E2 &
\begin{tabular}[c]{@{}l@{}} A judge must certify that {\color{red}\textbf{\textit{he}}} has familiarized {\color{red}\textbf{\textit{himself}}} with... \\ → All \textbf{\color{teal}\ul{judges}} must certify that \textbf{\color{teal}\ul{they}} have familiarized \textbf{\color{teal}\ul{themselves}} with...\end{tabular} \\
\hline
\multicolumn{3}{l}{\textbf{C. Relative and indefinite pronouns}}\\
\hline
EN & E5 & \begin{tabular}[c]{@{}l@{}}If a staff member is not satisfied...,  {\color{red}\textbf{\textit{he}}} can ask for a rehearing. \\ → Any staff member {\color{teal}\textbf{\ul{who}}} is not satisfied... can ask for a rehearing\end{tabular} \\
\hline
IT &  I3 & \begin{tabular}[c]{@{}l@{}}L’assicurazione... è a carico {\color{red}\textbf{\textit{del fruitore}}} [of the user]. \\ → a carico di \textbf{\color{teal}\ul{chi fruisce}} [of who uses].\end{tabular} \\
\hline
\multicolumn{3}{l}{\textbf{D. Collective and Role nouns}}\\
\hline
EN & § & \begin{tabular}[c]{@{}l@{}}Please contact one of the \color{red}\textbf{\textit{waiters}}. \\ → Please contact our \color{teal}\textbf{\ul{staff}}. \end{tabular} \\
\hline
IT & I3 & \begin{tabular}[c]{@{}l@{}}Il palazzo ospita gli studi {\color{red}\textbf{\textit{dei professori}}} [of the professors] di slavo. \\ → Il palazzo ospita gli studi {\color{teal}\textbf{\ul{del personale docente}}} [of the teaching staff] di slavo.\end{tabular} \\
\hline
\multicolumn{3}{l}{\textbf{E. Omission}}\\
\hline
EN & § & \begin{tabular}[c]{@{}l@{}}A person must reside... before {\color{red}\textbf{\textit{he}}} may apply for permanent residence. \\ → ...before {\color{teal}\textbf{\_\_}} applying for permanent residence.\end{tabular} \\
\hline
IT & I3 & \begin{tabular}[c]{@{}l@{}}Un’accurata compilazione facilita {\color{red}\textbf{\textit{allo studente}}} [to the student] diverse  \\ → ...facilita {{\color{teal}\textbf{\_\_}}} diverse operazioni. \end{tabular} \\
\hline
\multicolumn{3}{l}{\textbf{F. Repetition}}\\
\hline
EN & E3 & \begin{tabular}[c]{@{}l@{}}A manager may apply... if permission has been granted by {\color{red}\textbf{\textit{his}}} institution. \\ → ...if permission has been granted by {\color{teal}\textbf{\ul{that manager}}}’s institution.\end{tabular} \\
\hline
\multicolumn{3}{l}{\textbf{G. Passive voice}}\\
\hline
EN & E5 & \begin{tabular}[c]{@{}l@{}}Each action officer must send {\color{red}\textbf{\textit{his}}} document. \\ → Documents {\color{teal}\textbf{\ul{must be sent}}}.\end{tabular} \\
\hline
IT & I1 & \begin{tabular}[c]{@{}l@{}}{\color{red}\textbf{\textit{Il}}} richiedente presenta la domanda [The applicant submits the application]. \\ → La domanda {\color{teal}\textbf{\ul{va presentata}}} [The application must be submitted].\end{tabular}  \\
\hline
\multicolumn{3}{l}{\textbf{H. Imperative forms}}\\
\hline
EN & E5 & \begin{tabular}[c]{@{}l@{}}Each staff member is requested to submit {\color{red}\textbf{\textit{his}}} information. \\ → Please {\color{teal}\ul{\textbf{submit}}} all information. \end{tabular} \\
\hline
IT & § & \begin{tabular}[c]{@{}l@{}}{\color{red}\textbf{\textit{Il cittadino}}} deve allegare [The citizen must attach] un documento. \\ → {\color{teal}\textbf{\ul{Allega}}} [Attach] un documento.\end{tabular} \\
\hline
\multicolumn{3}{l}{\textbf{I. Impersonal forms}}\\
\hline
IT & I15 & \begin{tabular}[c]{@{}l@{}}{\color{red}\textbf{\textit{Il candidato}}} decade [The candidate loses] dal diritto... \\ → {\color{teal}\textbf{\ul{Si decade}}} [*One loses] dal diritto... \end{tabular} \\
\hline

\end{tabularx}
\caption{Examples of neutralization strategies. In \textbf{{\color{red}\textit{red, italic}}} the generic masculine formulations; in \textbf{{\color{teal}\ul{green, underlined}}} the gender-neutralizations. 
Column 2 provides the reference  to the (E)nglish/(I)talian guidelines where each
example was found (E1,2,3,..). If no example was found for a specific strategy within the guidelines, but the strategy is nonetheless be applicable, we fabricated an example (indicated with §). If a strategy is not applicable in one language, the corresponding example was omitted.
}
\label{tab:strategies}
\end{table*}

\subsection{Neutralization Strategies}
\label{sec:what-how}

Moving on to the gender-neutralization strategies, here we discuss them through a multilingual perspective, focusing on their practical implications.
In Table \ref{tab:strategies}, we also offer a systematization that attempts to map strategies across English and Italian – except for highly language-specific solutions that are impossible to transfer.


Concerning \textbf{\textit{what should be neutralized}}, 
we identify that these documents tend to largely focus on a particular form of gender discrimination
: masculine generics.
Masculine generics have been historically employed in administrative/legal 
texts to briefly refer to the public at large (e.g., see example B, where \textit{he} refers to the whole occupational category of \textit{judges}, and the Italian \textit{il docente} [M], \textit{professor} for the full teaching body). In the same vein, stereotypical associations and androcentric forms are discouraged (e.g., see example A in English). 
Overall, these guidelines are mostly concerned with generic referents. As we will discuss in Section \ref{sec:desiderata}, however, there are also 
circumstances where avoiding gender marks is necessary, e.g., to avoid misgendering individuals.
Finally, and from a linguistic standpoint, we underscore that – as expected – English gender-inclusive strategies focus on the neutralization of pronouns (e.g., C, E), which are the main carrier of gender distinction in notional languages. 
Instead, the Italian guidelines prioritize the neutralization of nouns 
thus overlooking adjectives, pronouns, and verbs, which are subject to gender agreement too. 
Although the analyzed sentences are simple toy examples within an institutional genre, effective gender-inclusive solution should take into consideration the full range of gendered words in grammatical gender languages.

In light of the foregoing, we now delve into \textit{\textbf{how to avoid gender discrimination in language}}. As previously anticipated, these top-down guidelines advocate for the use of neutralization strategies that conform to standardized, institutional language, 
over innovative, uncertain forms. 
As shown in Table~\ref{tab:strategies}, neutral solutions can vary greatly, ranging from omissions (e.g., E), and simple replacements of single words with 
epicene or collective nouns (e.g., A, B, D), to more complex reformulations that involve structural changes at the sentence level (e.g., F, G, H, I). 
On the one hand, though elegant, nouns replacement might be limiting if other gender-marked words are present, and only allow for a partial neutralization, e.g., as in IT \textit{\textbf{\ul{Il}}} [M] \textit{\textbf{\ul{professore}}} [M] \textit{è} \textit{\textbf{tenut\ul{o}}} [M] \textit{a rispondere} (EN \textit{The professor must answer}) neutralized as \textit{\textbf{L’insegnante} è \textbf{tenut}}\textit{\textbf{\ul{o}}} [M]. Moreover, the contextual nature of synonymy makes the choice of gender-neutral alternatives strictly case-specific \cite{edmonds-hirst-2002-near}. 
When possible, however, the neutralization of short segments 
appears preferable as it makes the outcome more fluent, as opposed to more complex phrasings. This strategy is not always viable, though. Consider, for instance, the Italian term “\textit{figli\ul{o/a}}” (EN \textit{child}): in lack of epicene synonyms, neutralization would require verbose periphrases, e.g., IT \textit{minore a carico} (EN \textit{underage, dependent child}) or \textit{persona che si è concepita o adottata} (EN \textit{person who was conceived or adopted}). 
 
Neutralization strategies emerge as complex choices, to be carefully selected and weighted so as to preserve the effectiveness of communication and the acceptability of a text, i.e, features like fluency, style. Such choices, of course, highly depend on various constraints (e.g., register, length, context of use). Therefore, when adopting inclusive language, it is crucial to consider the possible trade-off between neutrality and the overall acceptability of the text where it is implemented.
Moreover, as previously discussed, the feasibility and efficacy of adopting neutral strategies 
heavily depends on the context and the content of the source text. Therefore, such strategies are expected to be particularly pertinent in certain contexts, such as the administrative-institutional domain – to which, it is worth noting, most monolingual guidelines belong. Different and less formal textual styles and contexts could present harder challenges in performing GNT because of the higher heterogeneity of their texts, where the strategies presented above might prove inapplicable or inappropriate
. For instance, consider the translation of the simple, colloquial sentence EN \textit{I have never been there} into Italian (\textit{Non sono mai \ul{stato/stata} lì}): none of the strategies in Table \ref{tab:strategies} is applicable here. However, compared to institutional and administrative communication, colloquial contexts tend to have greater tolerance to 
creative translations (e.g., IT \textit{Non ho mai messo piede lì} -- literally, EN \textit{I have never set foot there}). Whether the system should resort to similar (or other) devices when straightforward solutions such as the strategies discussed above are not applicable is a decision that should be taken into account when building inclusive MT systems.


\section{Desiderata for GNT in MT}
\label{sec:desiderata}

In light of both the insights that emerged in Section \ref{sec:guidelines}, we now specifically address the use case of GNT, which allows MT systems to avoid discriminatory practices while conforming to standard linguistic forms.
Specifically, \textbf{we define GNT as the task of automatically translating from one language into another without marking the gender of human referents in the target.}
For example, given the English sentence \textit{Your \ul{neighbors} will thank you}, an inclusive MT system is required to translate \textit{Il vostro vicinato}\footnote{While the word \ul{\textit{vicinato}} is formally masculine, as a collective noun 
it is conceived as conceptually neutral.} \textit{vi ringrazierà}, as opposed to \textit{I vostri \ul{vicini} vi ringrazieranno}, which features a masculine generic. 

One crucial aspect of GNT is to determine when it should be performed, namely, when the marking of gender should be avoided or preferred. 
To this aim, and informed by our analysis of the existing guidelines, we devise three main desiderata to obtain a gender-neutral MT output, with specific examples in Table \ref{examples}.

\paragraph{\textbf{D1. Gender should not be expressed in the output translation when it cannot be properly assumed in the source.}} 
An inclusive MT system is expected to perform a gender-neutral translation in the target language when the gender of the referent(s) cannot be properly assigned from the source. This scenario is quite frequent when translating from a notional gender language into a grammatical gender one, because of the gap in gender expression we discussed in Section \ref{sec:genderlanguage}.
In these cases a gender-neutral translation refrains from any gratuitous assumptions, thus avoiding expressions which may: \textit{i)} misgender a specific referent 
(Example 1); \textit{ii)} exclude a social group, such as in the case of masculine generics 
(Ex. 2); \textit{iii)} foster stereotypical associations 
(Ex. 3); adopt "androcentric" expressions 
(Ex. 4).

\begin{table*}[h]

\centering
\small
\renewcommand{\arraystretch}{1.2}
\begin{tabular}{lll}

\textbf{(1)} & 
\begin{tabular}[c]{@{}l@{}}EN \\  IT \\ GNT \end{tabular} & 
\begin{tabular}[c]{@{}l@{}}I refuse to give up on \textbf{a single student} in my class. \\ Mi rifiuto di lasciare indietro \textbf{\color{red}\textit{un solo} \color{teal}\ul{studente}} nella mia classe. \\ Mi rifiuto di lasciare indietro \textbf{\color{teal}\ul{qualsiasi studente}} nella mia classe.\end{tabular} 
\\ \hline
\textbf{(2)} & 
\begin{tabular}[c]{@{}l@{}}EN \\  IT \\ GNT \end{tabular} & 
\begin{tabular}[c]{@{}l@{}}\textbf{A lot of innovative teachers} began bringing comics... \\ \textbf{\color{red}\textit{Molti} \color{teal}\ul{insegnanti} \color{red}\textit{innovativi}} iniziarono a portare i fumetti... \\ \textbf{\color{teal}\ul{Un gran numero di insegnanti all'avanguardia}} iniziarono a portare i fumetti...\end{tabular}
\\ \hline
\textbf{(3)} & 
\begin{tabular}[c]{@{}l@{}}EN \\  IT \\ GNT \end{tabular} & 
\begin{tabular}[c]{@{}l@{}}We train \textbf{nurses} to do it, and they use local anesthetics. \\ Formiamo \textbf{\color{red}\textit{le infermiere}} a farlo, e loro usano anestetici locali. \\ Formiamo \textbf{\color{teal}\ul{il personale infermieristico}} a farlo, e loro usano anestetici locali.\end{tabular}
\\ \hline
\textbf{(4)} & 
\begin{tabular}[c]{@{}l@{}}EN \\  IT \\ GNT \end{tabular} & 
\begin{tabular}[c]{@{}l@{}}Vehicles may only proceed \textbf{at walking pace}. \\ I veicoli possono procedere solo \textbf{\color{red}\textit{a passo d’uomo}}. \\ I veicoli possono procedere solo \textbf{\color{teal}\ul{a passo di persona}}.\end{tabular}
\\ \hline
\textbf{(5)} & 
\begin{tabular}[c]{@{}l@{}}EN \\  IT \end{tabular} & 
\begin{tabular}[c]{@{}l@{}}Even \textbf{the lead singer herself} abandoned the project. \\ Persino \textbf{\color{teal}{\ul{la stessa cantante solista}}} ha abbandonato il progetto. \end{tabular}
\\ \hline
\textbf{(6)} & 
\begin{tabular}[c]{@{}l@{}}EN \\  IT \end{tabular} & 
\begin{tabular}[c]{@{}l@{}}It affects one to two percent of the population, more commonly \textbf{men}. \\ Riguarda dall’uno al due percento della popolazione, ed è più comune negli \textbf{\color{red}{\ul{uomini}}}. \end{tabular}
\\ \hline
\textbf{(7)} & 
\begin{tabular}[c]{@{}l@{}}EN \\  IT \\ GNT \end{tabular} & 
\begin{tabular}[c]{@{}l@{}}Earth was pristine before \textbf{men} appeared. \\ La Terra era incontaminata prima della comparsa \textbf{\textit{\color{red}degli uomini}}. \\ La Terra era incontaminata prima della comparsa \textbf{\color{teal}\ul{degli esseri umani}}.\end{tabular}
\\ \hline
\textbf{(8)} & 
\begin{tabular}[c]{@{}l@{}}EN \\  IT \\ GNT \end{tabular} & 
\begin{tabular}[c]{@{}l@{}}The \textbf{fishermen} were so upset about not having enough fish to catch that... \\ I \textbf{\color{red}\textit{pescatori}} erano così \textbf{\color{red}\textit{disperati}} per la mancanza di pesce da pescare che... \\ \textbf{\color{teal}\ul{Le persone che pescavano}} erano così \textbf{\color{teal}\ul{disperate}} per la mancanza di pesce da pescare che... \end{tabular}
\\ \hline
\textbf{(9)} & 
\begin{tabular}[c]{@{}l@{}}EN \\  IT \end{tabular} & 
\begin{tabular}[c]{@{}l@{}}Now when I was a \textbf{freshman} in college, I took my first biology class. \\ Quando ero \textbf{\color{red}\ul{uno}} studente al primo anno di università, seguii il mio primo corso di biologia. \end{tabular} 
 \\

\end{tabular}

\caption{Examples for D1-3. We mark binary gender-marked expressions in \textbf{\color{red}\textit{red}}, and in \textbf{\color{teal}\ul{green}} those that are neutral.}
\label{examples}

\end{table*}

\paragraph{\textbf{D2. Proper expressions of gender should be generated in the output translation if they are (indirectly) expressed in the source.}}

The gender of some entities can be sometimes inferred through linguistic elements, which we may define as “gender cues”. For example, in English, gender cues are 3rd person pronouns (\textit{he/him/his}, \textit{she/her/hers}), terms of address (e.g., \textit{Mr./Mrs/Ms.}), gender-specific nouns (e.g, \textit{boy, lady, lord, wife}).
The presence of gender cues is crucial in determining whether a GNT is required or not.
In 
(Ex. 5), the pronoun \textit{herself} unequivocally identifies the referent as feminine.
First names, surnames, or even nicknames, however, should not be included among these cues for several reasons. First names can hardly be considered a reliable index of someone’s gender identity \cite{lauscher-etal-2022-welcome}. Even in the attempt of any binary correlation, names and nicknames are highly ambiguous across genders and cultures (e.g., \textit{Andrea}, which is typically masculine in Italian, but feminine in German). 
In addition, referents’ gender could be known also through non-textual elements, such as explicit external information about who is speaking, which is sometimes provided to the translators. In all these cases, gender expressions are preferable in the translation.

\paragraph{\textbf{D3. Masculine generics should not be propagated from the source language to the output translation.}}
In spite of the seemingly straightforward definition of gender cues in D2, their recognition might not be clear-cut. This is the case of masculine generics used in the source, whose distinction from an actual gender cue might be equivocal. Hence, a MT system should be brought to carefully consider every information, in particular the word \textit{man} along with its derivations and compounds so as to understand if they are used properly. For instance, to explicitly refer to the masculine gender group as a whole 
(Ex. 6), where a neutralization would effectively compromise the meaning of the sentence. On the contrary, when they are used to refer to the totality of human beings 
(Ex. 7), or to entire categories of mixed-gender people through terms such as \textit{fishermen} 
(Ex. 8), thus effectively functioning as masculine generics, they should be translated with neutral forms in the output. As there is not always a clear-cut distinction between a masculine generic and a masculine term used to refer to an actual masculine referent, and given the short context window within which MT systems operate, ambiguous cases can occur rather frequently. In these cases, a GNT should be considered the safest option as it avoids the propagation of the potential masculine generic without compromising the meaning of the sentence.
Nonetheless, 
there is a specific case where gender cues ought to be considered as 
trustworthy; namely, in relation to the speaker as 1st person singular referent 
(Ex. 9). Based on the assumption that speakers deliberately choose the most appropriate expressions while talking about themselves, such a choice should be respected in the output translation.

In conclusion, we have outlined a set of three overarching desiderata towards the purpose of gender-inclusive MT. 
Such a scaffolding represents our proposed set of guiding principles to be applied towards the development of more inclusive MT models based on gender-neutral translation. 
In the next Section, we discuss the technical challenges of implementing such desiderata in MT.

\section{Challenges and Insights for a Gender-Neutral Machine Translation}
\label{sec:challenges}

The adoption of neutral forms in MT could be conceived as a condition to be met or not met
, without any intermediate nuance.
The efficacy with which the condition is satisfied, on the contrary, can be rather nuanced; 
for example, there might be alternative inclusive solutions which might be perceived as more elegant or semantically closer to the input text, and others that satisfy these conditions to a lesser extent.
Therefore, from a formal perspective, gender inclusivity can be likened to the concept of \textit{constraint} \cite{garbacea}.

As a constraint, it shows a multifaceted character, which makes it comparable to other types of well-known constraints adopted in automatic language generation \cite{garbacea}.
First, as seen in Section \ref{sec:guidelines}, gender inclusivity can be linguistically realized through both specific lexical forms and syntactic constructions. For this reason, it can be likened to \textit{lexical} and \textit{syntactic constraints}. 
Then, the requirement of producing automatic translations that are as readable and fluent as possible, which is not always easily guaranteed in the case of neutral reformulations, makes gender inclusivity analogous to \textit{utility constraints} (i.e. the criteria by which a text must exhibit characteristics such as coherence, comprehensibility, and faithfulness) \cite{utility}.
Nevertheless, gender inclusivity also summarizes the manifold challenges of the aforementioned constraints, thereby demonstrating a higher level of complexity. Below we illustrate the major challenges of satisfying such a multifaceted constraint, focusing on the dynamicity of the neutralization strategies (Section \ref{dynamicity}), the dearth of adequate training data and methods (Section \ref{training}), and the lack of evaluation procedures (Section \ref{evaluation}).

\subsection{Addressing the Dynamic Nature of Gender Inclusivity}
\label{dynamicity}

To prevent unintended neutralizations, it is not always advisable to ensure GNT at all times (see Section \ref{sec:desiderata}).
This condition makes neutral translation a ``dynamic constraint", requiring MT systems to determine when to apply GNT. 
This ability, however, may be challenging to acquire, especially when gender cues are available outside the limited sentence context (e.g., \textit{He was talking with a young \ul{man}. Only later I realized that \ul{this person} was a \ul{professor}}).
This presents a problem for current state-of-the-art MT systems, which work at the sentence level, i.e., by translating each sentence in isolation.

 
Alternative solutions that account for larger textual context in translation \cite{lopes-etal-2020-document} might be more apt to decide when performing neutral translations.
For example, the design of MT models that translate beyond the sentence level ought to be considered. Translating sentences in a wider context, indeed, has proven crucial for correctly handling discourse cohesion \cite{bawden-etal-2018-evaluating}, and was shown to a certain extent beneficial to mitigate gender bias \cite{basta-etal-2020-towards}. 
However, it remains occasionally dubious whether context provides a useful linguistically-motivated knowledge \cite{kim-etal-2019-document,li-etal-2020-multi-encoder}. Before venturing into any document-level endeavor, it is thus recommended to verify whether there is a positive interpretable link between gender-neutral translation, context-informed MT,
and overall quality of the system.

Besides gender cues, explicit external information too may contribute to the disambiguation of gender in the source sentence, thus guiding the neutral translation. 
For instance, speakers' metadata can be supplied in the form of tags, either at the word level \cite{stafanovics-etal-2020-mitigating}
or at the sentence level \cite{vanmassenhove-etal-2018-getting,basta-etal-2020-towards}. 
Such prior knowledge, therefore, can also provide assistance in addressing the dynamic nature of gender inclusivity.

\subsection{Constraining MT systems towards GNT}
\label{training}

Future GNT-capable models are expected to learn to map words referring to human referents to corresponding neutral translations in order to satisfy the desiderata D1-3. Ideally, these models should be able to learn this mapping 
based on extensive training sets that include pairs of sentences with gender-neutral translations.
To the best of our knowledge, however, 
training data that consistently have neutral forms in the target side (with a grammatical gender language as target) is lacking.
It is necessary, then, to think of training methods that can overcome this lack of data, for example by taking inspiration from methods already applied to MT to satisfy other types of constraints.

Although various strategies have been proposed to make systems meet constraints (for an overview see \cite{garbacea}), it is crucial to evaluate which ones are applicable to the objective of gender inclusivity and how they can be adapted accordingly.
The most straightforward method, for example, is to make the constraint explicit to the model directly in the input data.
In the case of lexical constraints, this has been done by appending the constraint in the form of a target word or lemma to the source input so as to encourage the model to copy it in the output \cite[\textit{inter alia}]{dinu-etal-2019-training,song-etal-2019-code,chen-lexicalconstraints}.
However, this approach is designed to work mainly at word-level, hence it would not be suitable when neutralization should involve several segments of the sentence.
Moreover, this method requires bilingual dictionaries to map source words to target words.
For gender inclusivity, however, such terminologies are not available, yet.
Upon their creation, this technique could be taken into account when dealing with neutral source words which may be suitably translated with target epicene words.

Another line of solutions consists in restricting the search space at decoding time to sequences that contain the pre-defined constraints, such as specific words or phrases in the lexically-constrained MT.
For example, Hokamp \& Liu \shortcite{hokamp-liu-2017-lexically} and Post \& Vilar \shortcite{post-vilar-2018-fast} proposed modified versions of the beam search, 
which ensure that the translation hypotheses have met all the constraints before concluding the search.
Similarly, Saunders \& Byrne \shortcite{saunders-byrne-2020-reducing} and Saunders et al. \shortcite{saunders-etal-2022-first} designed a constrained beam search pass to improve gender diversity -- but for masculine/feminine forms only -- in the \textit{n}-best list by producing synthetic gendered alternatives of the original best hypothesis.
Alternatively, some approaches were proposed to re-rank the \textit{n}-best hypotheses according to additional scores, which informed whether or to which extent the constraints were satisfied, like in the dubbing-optimized MT \cite{saboo-baumann-2019-integration} or in gender-specific translations \cite{saunders-etal-2022-first}.
Decoding and re-ranking methods, however, may also entail outputs of lower quality \cite{saboo-baumann-2019-integration,chousa-morishita-2021-input}, due to the restriction of the search space and the trade-off between the need to satisfy the constraint and to faithfully reproduce the source text. Therefore, although such approaches may be a promising way to ensure gender inclusivity in automatic translations, their adoption too should be carefully evaluated.

\subsection{Evaluating Gender-Neutral Outputs}
\label{evaluation}

The lack of dedicated test sets and metrics 
prevents the possibility of determining whether systems are actually making any advancements towards the resolution of 
a given task.
In the case of gender-neutral MT, the benchmarks -- traditionally designed as parallel data for reference-based evaluations -- should comprise a range of source sentences aligned with target ones expressing either gender-marked or gender-neutral forms.
As a suitable starting point, the domain of such a test set 
could be 
based on the institutional/administrative texts, since the guidelines available for gender-inclusive language belong to this domain (see Section \ref{sec:guidelines}).
In addition to 
parallel data, 
specific protocols should also be designed to effectively evaluate whether the 
neutrality constraint has been satisfied. 

Typically, MT evaluation methods involve comparing the output with a reference and measuring the degree of overlap between 
n-grams \cite{papineni-2002-bleu,popovic-2015-chrf} 
or the distance between the generated sentence and the reference in terms of edit operations required to make them equal \cite{snover-etal-2006-study}. 
%
Some more sophisticated metrics take into account not only exact matches but also stems, synonyms, and paraphrases when comparing the MT output with the reference translation \cite{banerjee-lavie-2005-meteor}. 
Alternatively, neural metrics use models to predict the similarity between the output and reference (or even directly between the source and output) \cite{rei2020comet}.
%
Although metrics that do not rely solely on surface similarity may be more appropriate for evaluating gender neutrality,
it may be preferable to develop accuracy-like scores that isolate the evaluation of gender neutrality from the overall translation quality. 
This could involve annotating such expressions in the reference translation and attempting to match them, as done in MuST-SHE \cite{bentivogli-2020-gender}. In such cases, accuracy is 
determined through string matching between expressions in the reference and in the output. Hence, the risk of mismatch remains present, 
as automatic neutralizations may be difficult to detect in an evaluation pipeline based on a single reference and may require extensive manual analysis to be identified \cite{savoldi-lens}. 
Using multiple references \cite{qin-specia-2015-truly} that contain different neutral realizations to account for language variability could alleviate this difficulty.
Another option would be to calculate accuracy without exploiting reference translations, as designed in WinoMT \cite{stanovsky-etal-2019-evaluating}. 
In WinoMT the aim is to identify the gendered translation through word alignment with the source, determine its gender through a morphological analyzer, and then check whether 
it corresponds to that of the source. However, our scenario includes an additional challenge, as in grammatical gender languages gender-neutral expressions may carry a formal gender (e.g. \textit{la persona interessata} is a gender-neutral alternative of the masculine generic \textit{interessato}, but 
it is formally feminine). Thus morphological analysis may be problematic.

Overall, effectively evaluating whether the output of an MT system is gender-neutral or gender-marked presents several challenges. These challenges need to be addressed to develop an accurate approach that can overcome the limitations of overall translation quality metrics and account for the intrinsic variability of gender-neutral solutions.

\section{Conclusions}
\label{sec:conclusions}
As a promising route forward to counter 
gender bias, in this work we have taken the first steps towards the adoption of gender-inclusive language in MT, 
focusing on 
the use of neutral forms devoid of gender marking for an English-Italian setting.
To this aim, we 
reviewed various 
gender neutralization strategies presented in English and Italian 
guidelines for inclusivity, 
and 
outlined a definition of gender-neutral translation (GNT). 
Finally, we 
identified and discussed the technical challenges involved in implementing GNT in MT.

\section{Acknowledgements}
This work is part of the project ``Bias Mitigation and Gender Neutralization Techniques for Automatic Translation'', which is financially supported by an Amazon Research Award AWS AI grant.

Moreover, we acknowledge the support of the PNRR project FAIR - Future AI Research (PE00000013),  under the NRRP MUR program funded by the NextGenerationEU.



\bibliography{eamt23}
\bibliographystyle{eamt23}

\appendix
\section{Appendix}

\subsection{Guidelines}
\label{guidelines}

The following guidelines for gender-inclusive language were analyzed for this study:
\begin{itemize}
\item[E1]
United Nations Economic Commission for Western Asia, 2014 \\
\url{https://archive.unescwa.org/sites/www.unescwa.org/files/page_attachments/1400199_0.pdf}.
\item[E2]
United Nations, 2018 \\
\url{https://www.un.org/en/gender-inclusive-language/guidelines.shtml}.
\item[E3]
General Secretariat, Council of the European Union, 2018. \\
\url{https://www.consilium.europa.eu/media/35446/en_brochure-inclusive-communication-in-the-gsc.pdf}
\item[E4] European Parliament, 2018 \\
\url{https://www.europarl.europa.eu/cmsdata/187115/GNL_Guidelines_EN-original.pdf}
\item[E5] North Atlantic Treaty Organization, 2020 \\ \url{https://www.nato.int/nato_static_fl2014/assets/pictures/images_mfu/2021/5/pdf/210514-GIL-Manual_en.pdf} 
\item[E6] Australian Government, 2021 \\ \url{https://www.stylemanual.gov.au/accessible-and-inclusive-content/inclusive-language/gender-and-sexual-diversity}
\item[E7] University of Houston, 2022 \\ \url{https://www.uh.edu/marcom/guidelines-policies/inclusive-language/_files/inclusive-language-guide.pdf} 
\item[E8] Australian National University, n.a. \\ \url{https://services.anu.edu.au/human-resources/respect-inclusion/gender-inclusive-language} 
\item[E9] United Nations Women, n.a. \\ \url{https://authoring.prod.unwomen.org/sites/default/files/Headquarters/Attachments/Sections/Library/Gender-inclusive%20language/Guidelines-on-gender-inclusive-language-en.pdf} 
\item[E10] University of North Carolina at Chapel Hill, n.a. \\ \url{https://writingcenter.unc.edu/tips-and-tools/gender-inclusive-language/}
\item[E11] University of Pittsburgh, n.a. \\ \url{https://www.gsws.pitt.edu/resources/faculty-resources/gender-inclusive-non-sexist-language-guidelines-and-resources}
\item[E12] Royal Melbourne Institute of Technology, n.a. \\ \url{https://www.rmit.edu.au/content/dam/rmit/au/en/students/documents/services-support/lgbtiq/guide-inclusive-language.pdf}
\item[E13] California State University San Marcos, n.a. \\ \url{https://www.csusm.edu/ipa/surveys/inclusive-language-guidelines.html}
\item[E14] University of Otago, n.a. \\ \url{https://www.otago.ac.nz/humanresources/working-at-otago/equity/inclusive-language/index.html}
\item[E15] The University of Texas at Austin, n.a. \\ \url{https://intranet.dellmed.utexas.edu/public/inclusive-language-guidelines}
\item[I1] Cancelleria Federale Svizzera, 2012 \\ \url{https://www.bk.admin.ch/dam/bk/it/dokumente/sprachdienste/Sprachdienst_it/02/objekt_40366.pdf.download.pdf/guida_al_pari_trattamentolinguisticodidonnaeuomo.pdf}
\item[I2] Università di Torino, 2015 \\ \url{https://www.unito.it/sites/default/files/linee_guida_approccio_genere.pdf}
\item[I3] Università degli Studi di Padova, 2017 \\ \url{https://www.unipd.it/sites/unipd.it/files/2017/Generi%20e%20linguaggi.pdf}
\item[I4] Segretariato Generale, Consiglio dell'Unione Europea, 2018 \\ \url{https://www.consilium.europa.eu/it/documents-publications/publications/inclusive-comm-gsc/}
\item[I5] Parlamento Europeo, 2018 \\ \url{https://www.europarl.europa.eu/cmsdata/187102/GNL_Guidelines_IT-original.pdf}
\item[I6] Università degli Studi di Verona, 2020 \\ \url{https://docs.univr.it/documenti/Documento/allegati/allegati044384.pdf}
\item[I7] Università di Bologna, 2020 \\ \url{https://www.unibo.it/it/allegati/linee-guida-per-la-visibilita-del-genere-nella-comunicazione-istituzionale-dell2019universita-di-bologna/@@download/file/Linee-Guida-Genere-2020.pdf}
\item[I8] Università degli Studi dell'Aquila, 2020 \\ \url{https://www.univaq.it/include/utilities/blob.php?item=file&table=allegato&id=4925}
\item[I9] Università di Siena, 2021 \\ \url{https://www.unisi.it/sites/default/files/allegatiparagrafo/LINEE_GUIDA_Linguaggi_e_Generi.pdf}
\item[I10] Istituto Universitario Federale per la Formazione Professionale, 2021 \\ \url{https://www.suffp.swiss/sites/default/files/guida_per_un_linguaggio_inclusivo_20200610.pdf}
\item[I11] Università della Calabria, 2021\\
\url{https://www2.unical.it/portale/strutture/dipartimenti_240/fisica/pariopportunita/Linee%20guida%20Linguaggio%20di%20genere_15%20marzo%2021.pdf}
\item[I12] Università degli Studi di Milano, 2021 \\ \url{https://www.unimi.it/sites/default/files/2021-12/Vademecumlinguaggio%20di%20genere_Universit%C3%A0%20degli%20Studi%20di%20Milano.pdf}
\item[I13] Università Mediterranea di Reggio Calabria, n.a. \\ \url{https://www.unirc.it/documentazione/media/files/ateneo/pari_opportunita/File_allegato_2.pdf}
\item[I14] Università di Trento, n.a. \\ \url{https://www.unitn.it/alfresco/download/workspace/SpacesStore/1185b2b5-dcfe-48ef-882b-e7042fe4ff1a/documentolinguaggio29mar%20(1).pdf}
\item[I15] Università di Ferrara, n.a. \\\url{https://drive.google.com/file/d/1P5Eq2jjoJtTjXGEV7TzyM4XJTcV2PRyp/view}

\end{itemize}



\end{document}